\documentclass{article}

\usepackage{PRIMEarxiv}

\usepackage[utf8]{inputenc} 
\usepackage[T1]{fontenc}    
\usepackage{hyperref}       
\usepackage{url}            
\usepackage{booktabs}       
\usepackage{amsfonts}       
\usepackage{nicefrac}       
\usepackage{microtype}      
\usepackage{lipsum}
\usepackage{fancyhdr}       
\usepackage{graphicx}       
\graphicspath{{media/}}     

\usepackage{amsmath}
\usepackage{subcaption}

\usepackage{booktabs}
\usepackage{threeparttable}

\title{Model discovery for dynamical systems with complex-valued product units
}

\author{
  Martin Brückmann, Babette Dellen, Uwe Jaekel \\
  Department of Mathematics, Informatics, and Technology, RheinAhrCampus Remagen \\
  University of Applied Sciences Koblenz \\
  Joseph-Rovan-Allee 2, 53424 Remagen, Germany\\
  \texttt{\{brueckmann, dellen, jaekel\}@hs-koblenz.de} \\
}

\begin{document}
\maketitle

\begin{abstract}
  Discovering the governing equations of a dynamical system from observed
  trajectories provides deeper insight into its structure than mere prediction
  of future states. We present a data-driven approach to model discovery based
  on complex-valued product-unit networks, in which each unit represents a
  complex monomial and the network output is a sparse linear combination of such
  monomials. In contrast to established library-based methods such as SINDy, our
  approach does not require a predefined set of candidate functions: the
  relevant monomials, including those with fractional or negative exponents, are
  learned directly from data. Across four chaotic benchmark systems (Lorenz63, Lorenz84, the Four-Wing attractor, and a fractional variant of Lorenz63), we recover the exact governing equations in 90\% of trials for the first three systems, and in 70–90\% of trials for the fractional case, using at least 3000 training points. Applied to real-world
  human-gait accelerometer signals, the model produced stable trajectories with bounded prediction errors, corresponding to an RMSE of approximately 12-14\% of the signal amplitude range over a test horizon three times longer than the training interval, demonstrating its potential for high-dimensional systems in which analytic equations are unavailable.
\end{abstract}

\section{Introduction}
It is well known that highly complex patterns and even chaos in dynamical
systems can emerge from comparatively simple governing equations. For example,
the chaotic dynamics of the Lorenz attractor
\cite{Lorenz1963,PhysRevLett.96.034104}, originally serving as a model for
weather phenomena, arises from three sparse nonlinear polynomial equations,
consisting of monomials that couple the main variables of the system by
multiplication. The recovery of the accurate algebraic structure of the leading
monomials driving the dynamical system is highly valuable for gaining a better
understanding of the properties of the system, such as dynamic stability and
interaction of variables, as well as further theoretical insights. Examples of
real-world data originating from dynamical systems are physiological time series,
such as human-gait accelerometer data, or EEG \cite{Koliqi2024} and ECG
\cite{Nayak2018} data, but also machine vibrations in predictive maintenance
\cite{Ng2015}.
A major challenge in identifying the relevant monomials, or more generally terms
with fractional or negative exponents, for high-dimensional systems is the vast,
potentially infinite, number of candidate terms to be explored. Since this
number grows exponentially with the input dimension, exhaustive search quickly
becomes intractable.

In recent years, several neural-network-based approaches have been proposed to
learn sparse polynomial representations from data that have shown promising
results. First-order input couplings were implemented for a reservoir-computing
approach applied to the problem of forecasting the behavior of the Lorenz system
\cite{gauthier2021}. Higher-order couplings have also been used in the SINDy
approach to model nonlinear dynamical systems \cite{Brunton2016a}. Using a
preset candidate library of monomials, the active candidates are determined
using sparse regression.

Wang and Li (2024) have recently proposed the dynamical-system deep-learning
(DSDL) model to predict chaotic time series, which is based on a state-space
reconstruction (SSR) technique and can, unlike many traditional deep-learning
approaches, also infer important variables and high-degree interactions of the
system \cite{Wang2024}. Their architecture was shown to be effective with noisy
data \cite{Wu2025} and also delivered promising results for partial
differential equations \cite{Li2025}.

Another notable approach is symbolic regression, which aims to find short
mathematical expressions that describe the underlying data, typically making use
of genetic programming or the usually faster FFX algorithm. The latter utilizes
a large candidate set of combinations of nonlinear functions to subsequently
collect Pareto optimal expressions that trade off error and the number of terms
in the representation \cite{McConaghy2011, Quade2016}.

\begin{figure*}[tbhp]
\centering
\includegraphics[width=0.9\textwidth]{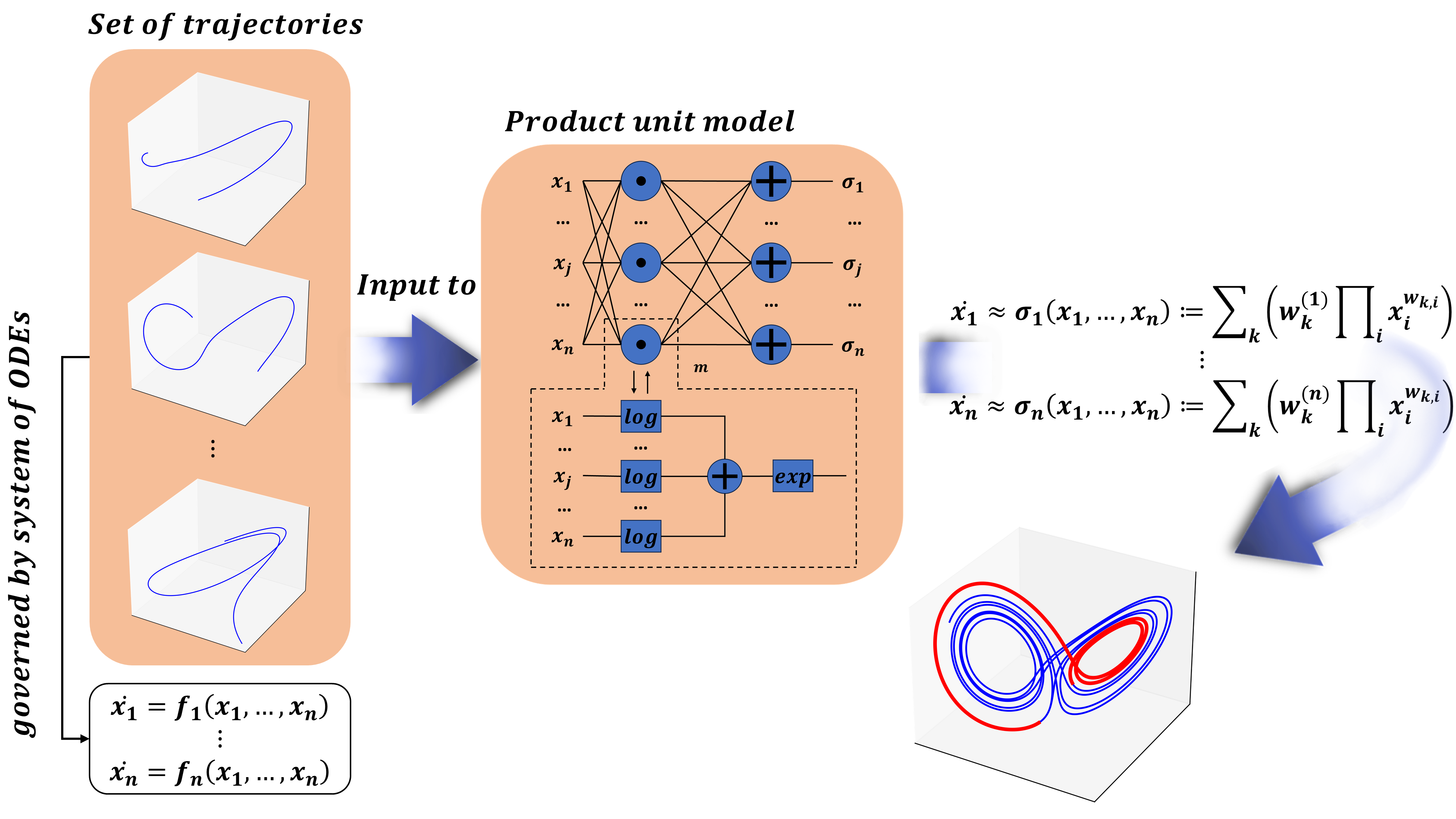}
\caption{General approach for systems following differential equations: We
  assume knowledge of a time series of training trajectories in phase space,
  consisting of points and the numerical values of their time derivatives. If
  values for the derivatives are not directly available, we assume that they can
  be approximated numerically. The data are assumed to be subject to a system of
  ordinary differential equations. By comparing evaluations of the true
  equations at these points with derivatives obtained from the model, the
  product-unit model fits the inputs to linear combinations of monomials to
  approximate the original equations, i.e., we learn the right-hand side (RHS)
  of equations $\dot x_j$. Furthermore, the derived equations permit prediction
  of future system states. The stability of trajectories generated from the
  model's equations can be assessed using the effective prediction time (EPT),
  which indicates the point at which the predicted trajectory diverges from the
  true trajectory.}
\label{fig:product_unit}
\end{figure*}

However, these approaches are limited because of their use of predefined
candidate functions or monomials, restricting the number of nonlinear function
classes that can be represented by the model, which is particularly severe for
high-dimensional problems. Therefore, we follow a different approach. Product
units are neural network units that can be trained to take the form of any
monomial or even algebraic terms containing fractional or negative exponents
\cite{durbin,leerink,fernandez2018time,dellenetal2019}, but which can be
difficult to train. Recently, it could be shown that complex-valued product unit
networks can successfully learn sparse approximations of real-valued functions
from real data and outperform classical multi-layer perceptrons in extrapolation
tasks \cite{DellenEtAl2024}. The complex-valued product-unit network operates
fully in the complex domain, meaning that its inputs, outputs, and weights are
complex-valued, which distinguishes it from previous approaches. Training of
this network could be achieved by standard gradient descent using a loss defined
for complex-valued outputs. In this work, we use complex-valued product unit
networks to learn the governing equations of dynamical systems directly from
observed trajectories of the system.


Here, we focus on dynamical systems with a moderate, but unknown number of
multiplicative couplings of fractional power laws in the dynamical variables.
Note that this includes the possibility of approximating other non-linear
couplings in terms of Taylor- or Laurent expansions. A major challenge in the
identification of the dominant monomial terms lies in the large number
of possible couplings, which grows exponentially with the number of
variables. Since the required leading terms of the sparse polynomial as well as
potentially fractional exponents in our approach are learned from the data
directly without using a candidate library, the problem of complexity-of-choice
is avoided.

To determine its aptitude in extracting the governing equations of a dynamical
system from state trajectories, the proposed method is tested on some standard
benchmark problems, i.e., the Lorenz63 attractor, Lorenz84 and a Four Wing
system. Furthermore, a modified instance of Lorenz63 is investigated, which now
includes a fractional exponent in the third equation. We finally apply the
product-unit model to real-world human-gait patterns in order to learn a
mathematical representation of human gait.



\section{Methods}
This section introduces the complex-valued product-unit model, the benchmark
dynamical systems used for evaluation, and evaluation measures. A general
overview of the approach can be seen in Fig.~\ref{fig:product_unit}: The model
is trained on a set of simulated trajectories (discrete points and their
derivatives) to learn the precise terms of the underlying differential
equations. Subsequently, the identified equations and the stability of their
predictions over time are assessed.

\subsection{Complex-Valued Product-Unit-Network Model}\label{section:model}
The model consists of nonlinear units that compute products of their inputs
according to $\prod_{i=1}^{n} x_i^{w_{i}}$ with complex-valued weights $w_i$
\cite{DellenEtAl2024}. In practice, these product units are implemented by
concatenating summation units with logarithmic and exponential activation
functions defined in the complex domain (Fig.~\ref{fig:product_unit}), such that
\begin{equation}\label{eq:1}
\prod_{i=1}^{n} x_i^{w_{i}} = \exp\!\left(\sum_{i=1}^{n} w_i \log x_i\right).
\end{equation}
The complex-valued formulation enables the model to handle negative and complex
inputs naturally. However, inputs cannot be zero, but this constraint rarely
poses practical limitations \cite{DellenEtAl2024}.

The resulting complex-valued units are subsequently supplied to a summation
layer, so that an equation is represented as a linear combination of
(generalized) monomials by
\begin{equation}\label{eq:2}
\hat f(x_1, \ldots, x_n) := \sum_{k=1}^{m} c_k \prod_{i=1}^{n} x_i^{w_{k,i}},
\end{equation}
where $c_k \in \mathbb{C}$ is the coefficient of the $k$-th monomial and $w_{k,i} \in \mathbb{C}$ is the exponent of variable $x_i$ in that monomial, following \cite{DellenEtAl2024}. The output of this
function is a complex number and both the real and the imaginary part are
supplied to the loss function.

For a system with multiple equations, we use a product-unit model that realizes
one sum per equation. For a set of three ordinary differential equations, as it
is the case for the benchmark problems (Sec.~\ref{section:choice}), we learn the
right-hand side (RHS) of the equations
\begin{align}\label{eq:3}
  \dot{x} &= f_x(x,y,z),\nonumber\\
  \dot{y} &= f_y(x,y,z),\nonumber\\
  \dot{z} &= f_z(x,y,z),
\end{align}
directly, so that each function $f_{v \in \{x,y,z\}}$ is represented by a
weighted sum of product units, denoted $\hat f_v$. For the real-world human gait
signals considered in this work, three-dimensional acceleration measurements
likewise allow the dynamics to be captured by three equations. However, in the
latter case we predict the future directly from past states.

For example, $f_{x}(x,y,z) = 5x^2y^3 + xz^{0.1} - 4$ can be rewritten in a
normalized form by expressing each term as a product of powers of the variables
$x, y$ and $z$:
\[
5x^2y^3z^0 + 1x^1y^0z^{0.1} + (-4)x^0y^0z^0.
\]
Thus, the formula can be represented by three product units.

Equations of dynamical systems may share monomials, for instance the term
$xy$ appears in both $\dot y$ and $\dot z$ of the Lorenz84 system. To
accommodate such shared terms, we allow the product-unit layer to be fully
connected with the summation-unit layer, leading to the mathematical model
given by Eq.~\ref{eq:2}.

The model is implemented in PyTorch \cite{NEURIPS2019_bdbca288}, using the formulation in Eq.~\ref{eq:1}, where each product unit is implemented as a complex-valued linear layer with bias in the logarithmic domain, followed by an exponential activation.

\subsection{Benchmark systems}\label{section:choice}
For benchmarking, we simulate the trajectories of several three-dimensional
dynamical systems using numerical methods. Therefore, we have access to
numerical approximations of the state variables and their time derivatives. From
these discrete observations, our goal is to recover the underlying differential
equations that govern the system. We then compare the parameters and structure
of the discovered model with the true system. Using this model, we approximate
trajectory predictions using numerical methods.

We first consider three known examples of chaotic dynamical systems of
first-order ordinary differential equations to show the general applicability of
the model in this regard: The Lorenz63 system \cite{Lorenz1963}, as well as
Lorenz84 \cite{Vannitsem2002}, and a Four-Wing chaotic attractor introduced by
Wang et al. (2009) \cite{Wang2009}. The systems vary in the minimum number of
product units required to represent their terms, which makes them good
candidates for benchmarking. The system Lorenz63 requires at least five product
units, Four Wing requires six, and Lorenz84 requires eight.
Furthermore, a modified version of Lorenz63 is considered which includes a term
with a fractional exponent.

The Lorenz63 system, originally proposed by E. N. Lorenz in 1963, consists of a
set of three coupled ordinary differential equations \cite{Lorenz1963},
\begin{align}
  \dot{x} &= \sigma (-x+y),\nonumber\\
  \dot{y} &= -xz + \rho x - y,\nonumber\\
  \dot{z} &= xy - \beta z.
\end{align}
Parameters are typically chosen as $\sigma = 10$, $\rho = 28$ and $\beta = 2.667$.

Lorenz84 is defined by the governing equations
\begin{align}
  \dot{x} &= -y^2 -z^2 -ax +aF,\nonumber\\
  \dot{y} &= xy -bxz -y + G,\nonumber\\
  \dot{z} &= bxy + xz - z,
\end{align}
where $a=0.25$, $b=6$, $F=16$, and $G=3$ \cite{Vannitsem2002}.

We further use a butterfly attractor (Four Wing model) \cite{Wang2009} with parameters $a=0.2$, $b=-0.01$, $c=1$, $d=-0.4$, $e=-1.0$, $f=-1$, and equations
\begin{align}
  \dot{x} &= ax + cyz,\nonumber\\
  \dot{y} &= bx + dy - xz,\nonumber\\
  \dot{z} &= ez + f xy.
\end{align}

\begin{figure}[htbp]
\centering
\includegraphics[width=.5\linewidth]{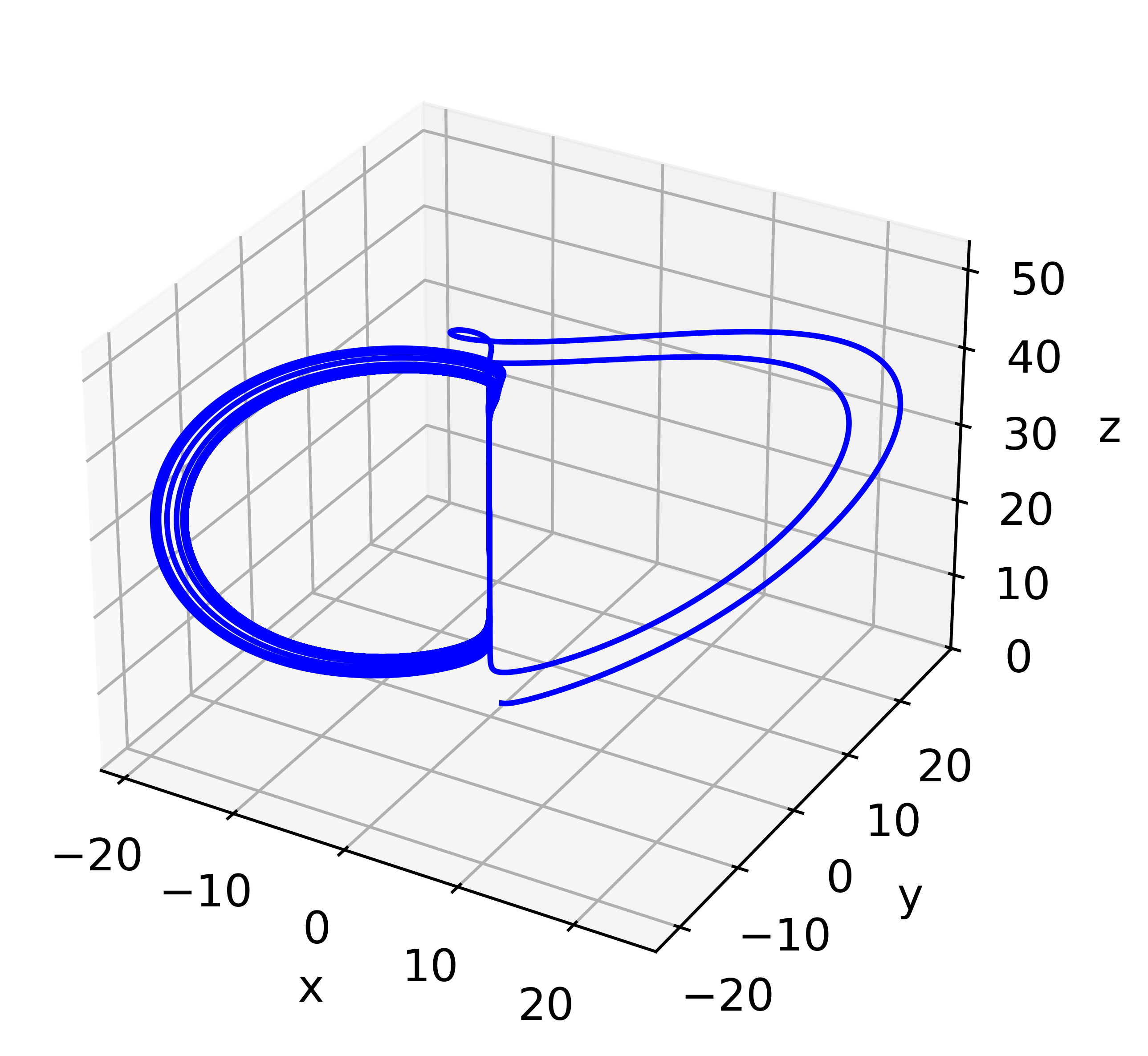}
\caption{A simulated trajectory derived from the modified Lorenz system Lorenz\_Fract. The structure consists of two loops of similar size, which are joined together by a narrow tube at their edges.}
\label{fig:LorenzMod}
\end{figure}

These systems contain only exponents strictly limited to nonnegative integer
values. Therefore, another system of ODEs is proposed that is a slightly
modified version of Lorenz63 but has a fractional exponent:
\begin{align}\label{eq:7}
  \dot{x} &= \sigma (-x+y),\nonumber\\
  \dot{y} &= -xz + \rho x - y,\nonumber\\
  \dot{z} &= xy - \beta z^{\eta}.
\end{align}
The parameters are set to $\sigma = 35$, $\rho = 28$, $\beta = 3$ and $\eta = 0.5$. The attractor is shown in Fig.~\ref{fig:LorenzMod} and is referred to as \textit{Lorenz\_Fract} in the following. Note that for negative $z$, $z^{\eta}$ with $\eta = 0.5$ becomes complex. Since our model operates in the complex domain, this poses no fundamental obstacle; however, for the numerical generation of training data we restrict trajectories to the positive orthant (see Sec.~\ref{section:data}) to ensure stable RK4 integration.


\subsection{Real-world data}\label{section:real}
Beyond these benchmark systems, we apply the product-unit model to accelerometer data from human gait. While previously differential equations were learned from state derivatives, here we use a time-delay
embedding approach, i.e., predicting the next point in a trajectory directly based on multiple previous states. This creates a high-dimensional input space, as the model must consider a sequence of past measurements to capture the system's time behavior.

Human gait exhibits close to periodic behavior and is roughly sinusoidal
\cite{Capecci2016, Yoo2002}. Since cosine and sine functions are solutions of
linear systems of ordinary differential equations (e.g.
$\dot x = -y; \dot y = x$), it is reasonable to assume that their governing
equations can be captured by our approach. However, because of the noise in the
data, spurious terms are to be expected.


\newpage
\subsection{Generation of Training Data}\label{section:data}
We consider four benchmark systems and real-world human gait data. For the benchmark systems, we simulate discrete time series from the system's equations to train the model, while the real-world data is directly measured using smartphone sensors.

\subsubsection{Phase-Space Trajectories from Benchmark Systems}
The state variables are governed by a system of ordinary differential equations as described in the previous section. The training data take the form of a discrete time series of multiple phase-space trajectories, consisting of both points $(x,y,z)$ and their corresponding time derivatives $\dot{x}=f_x(x,y,z), \dot{y}=f_y(x,y,z), \dot{z}=f_z(x,y,z)$.

\begin{figure}[htbp]
\centering
\includegraphics[width=.5\linewidth]{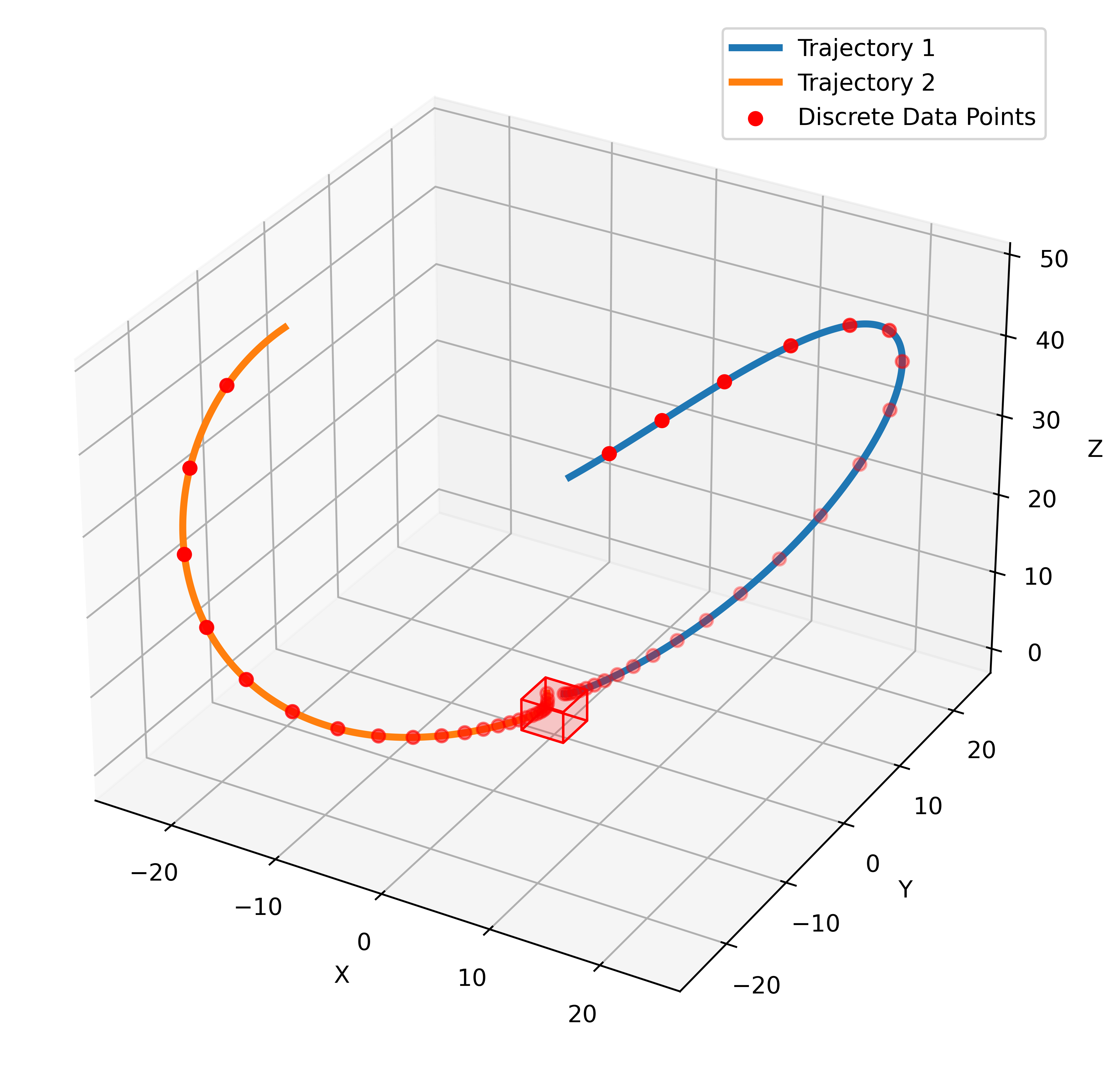}
\caption{Two example trajectories of the Lorenz63 system, each consisting of
  $500$ points. The initial coordinates are randomly generated within the
  interval $[-2,2]$, as indicated by the red cube. The subsequent $499$ points
  are computed using the fourth-order Runge-Kutta (RK4) method with a step size
  of $\Delta t = 0.001$. For clarity, only every 20th point along a trajectory
  is shown.}
\label{fig:trajExample}
\end{figure}

The points along the trajectories are generated according to the equations of the dynamical system. For each trajectory of size $m$, we begin by selecting an initial starting point $p := (x, y, z) \in \mathbb{R}^3$ with coordinates drawn uniformly at random from a fixed interval. The remaining $m-1$ points are computed sequentially by integrating the ODEs forward using the Runge-Kutta (RK4) method. An example is shown in Fig.~\ref{fig:trajExample}.
For the trajectories for the systems Lorenz63, Lorenz84, and Four Wing,  initial coordinates from $[-2, 2]$ and a step size of $\Delta t = 0.001$ are used. 

Furthermore, training of the system Lorenz\_Fract proved to be more challenging than training of other models when the range of initial values and the step size were kept the same. This could be due to the limited variation in the input. Trajectories start from a small interval and hence do not differ much at first. Moreover, the rate of divergence between two close trajectories is relatively low, as indicated by the largest Lyapunov exponent (see Sec.~\ref{section:modelevaluation}). Using the Python library \textit{Lyapynov} \cite{savary2025}, we estimate $\lambda_{\max}$ to be approximately $6 \times 10^{-5}$ for Lorenz\_Fract, which is several orders of magnitude smaller than the ones of the other attractors (see Table~\ref{table:Lyap}).

To increase variation in the training samples for Lorenz\_Fract, initial coordinates are drawn from a larger range, while the values of the variables $x$, $y$, and $z$ are constrained to non-negative numbers in the interval $[0, 20]$ to ensure numerical stability of the RK4 integration of the fractional power. The trained model itself remains fully complex-valued and is not subject to this restriction. In addition, a larger Runge--Kutta (RK4) step size of $\Delta t = 0.01$ is used in the construction of trajectories, allowing for a more thorough sampling of the attractor, although a smaller step size results in greater stability when integrating.

The product-unit model is then trained on multiple trajectories, with the number of trajectories ranging from $10$ to $100$ in increments of $10$, for a total of $1000$, $3000$, or $5000$ points, using gradient descent.

Because the trajectories are generated by simulation, their derivatives are also
available. In case of real (observed) data, either numerical derivatives have to
be computed first, or, as can be seen in the next section for the example of
gait analysis, the current data has to be predicted directly from previous data
points, as is commonly done in time-series prediction. However, additional
higher-order terms may arise from the described numerical procedures that are
not present in the true system due to approximations.
By selecting an appropriate loss function for complex-valued inputs, the model
can be trained using standard backpropagation and gradient descent algorithms.

\subsubsection{Gait Dynamics from Wearable Accelerometers}
The real-world signals are discrete observations of acceleration values along a
trajectory of a walking person. The product-unit model represents a
high-dimensional function that computes a current state from a sequence of
previous states. Thus, a training sample consists of several points, with the
subsequent point serving as the expected target, i.e.,
\begin{equation}
a_v(t) \approx \hat f_v(a(t-\psi_1), \ldots, a(t-\psi_m)), \quad v \in \{x,y,z\},
\end{equation}
where $a(t) = (a_x(t), a_y(t), a_z(t))$. To capture recurring
and periodic behavior, a sufficient number of time lags over an appropriate time
horizon is required. Time lags are chosen according to
\begin{equation}
\psi_k := \lfloor 500^{k/50} + k - 1\rfloor,~~~~~~k = 1,...,m,
\end{equation}
where $m := 50$. Consequently, the function lives in a $150$-dimensional domain
with the largest time lag equal to $\psi_{50}=549$. After the model has been
trained, points along a trajectory can be predicted sequentially. To improve
trajectory stability, here only the real component of previous points is used to
predict the next one, whereas the parameters of the model remain fully complex-valued.

To generate a signal, acceleration data are captured at a sampling rate of
approximately $200$ Hz using the \textit{phyphox} app \cite{Staacks2018} on an
Android device during a straight-line walk lasting $40$ seconds, corresponding
to $8000$ data points per signal. The first $10$ seconds ($2000$ points) of each signal are
used to train the model ($1451$ samples). In total, ten walking time series are
generated. For each time series, a model is trained and evaluated.

We used a 4th-order Butterworth low-pass filter with a cutoff frequency at $15$
Hz to remove noise from the time series before using them to train the models.

\subsection{Assessment of Model Performance for the Benchmark}\label{section:modelevaluation}
After training the model, the number of identified terms—both correct and
erroneous—is reported, with a margin of error allowed for any divergence from
expectation. Next, the stability of subsequent trajectory predictions
$\hat f_x(x, y, z), \hat f_y(x, y, z), \hat f_z(x, y, z)$ is evaluated.

\subsubsection{Counting the Number of Correctly Identified or Erroneous Terms}
To compare the terms discovered by the model with the true terms, similar terms
are merged and very small terms are excluded from the analysis.

After the last training iteration, the terms are merged if the exponents of the
corresponding variables differ by no more than $\epsilon = 0.1$. Specifically,
two terms $P_j$ and $P_k$ are merged if
\[
|w_{j,i} - w_{k,i}| \leq \epsilon \text{ for all } i \in \{1,\ldots,n\},
\]
where $w_{j,i}$ and $w_{k,i}$ denote the exponents of the variable $x_i$ in
$P_j$ and $P_k$, respectively. The exponents are then averaged and the
coefficients summed. For example, with three-dimensional input and assuming
similar exponents, two terms can be merged as
\begin{align}
    P_j + P_k &:= c_j x_1^{w_{j,1}} x_2^{w_{j,2}} x_3^{w_{j,3}} + c_k x_1^{w_{k,1}} x_2^{w_{k,2}} x_3^{w_{k,3}}\notag\\
    &\approx (c_j+c_k) x_1^{(w_{j,1} + w_{k,1})/2} x_2^{(w_{j,2} + w_{k,2})/2} x_3^{(w_{j,3}+w_{k,3})/2}.\notag
\end{align}
When counting correctly identified or erroneous components, we exclude terms
with absolute coefficients below $\delta = 10^{-3}$. Predicted terms where all
weights and coefficients match the true weights and coefficients with an
absolute difference of at most $\epsilon$ each are considered to be
\textit{correctly identified} terms, otherwise we call them \textit{erroneous}
terms.

\subsubsection{Measuring Forecasting Stability}
The robustness of the model can be evaluated with respect to its forecasting
ability. First, in order to remove remaining noise, rounding to the third
decimal place is applied to the generated system in advance. Then, to determine
the predictive power of the model over a certain period of time, the effective
prediction time (EPT) is calculated, as described by \cite{Wang2024}. From the
model equations, a trajectory is constructed, following the procedure described
in Sec.~\ref{section:data}. The initial coordinates are randomly selected from a
predefined interval: $[-4,4]$ for Lorenz63, Lorenz84, and Four Wing, and
$[0, 40]$ for Lorenz\_Fract. Subsequent points are computed using the
Runge-Kutta (RK4) method for both the true and the model system. We first
integrate forward for $5 \times 10^4$ steps by using the equations of the true
system. Then, the future trajectories are generated in the same way for the true
system and the predicted model and compared using the EPT.\@ The EPT corresponds
to a single point in time where a significant deviation from the expected
trajectory occurs. Time is represented as a discrete sequence of consecutive
steps, where $t \in \{1,...,m\}$ denotes the $t$-th point of a simulated
trajectory of size $m$.

\begin{figure*}[htbp]
    \centering
    \captionsetup[subfigure]{justification=raggedright, singlelinecheck=false}
    \begin{subfigure}[t]{0.62\textwidth}
        \caption{}
        \centering
        \includegraphics[width=11.05cm]{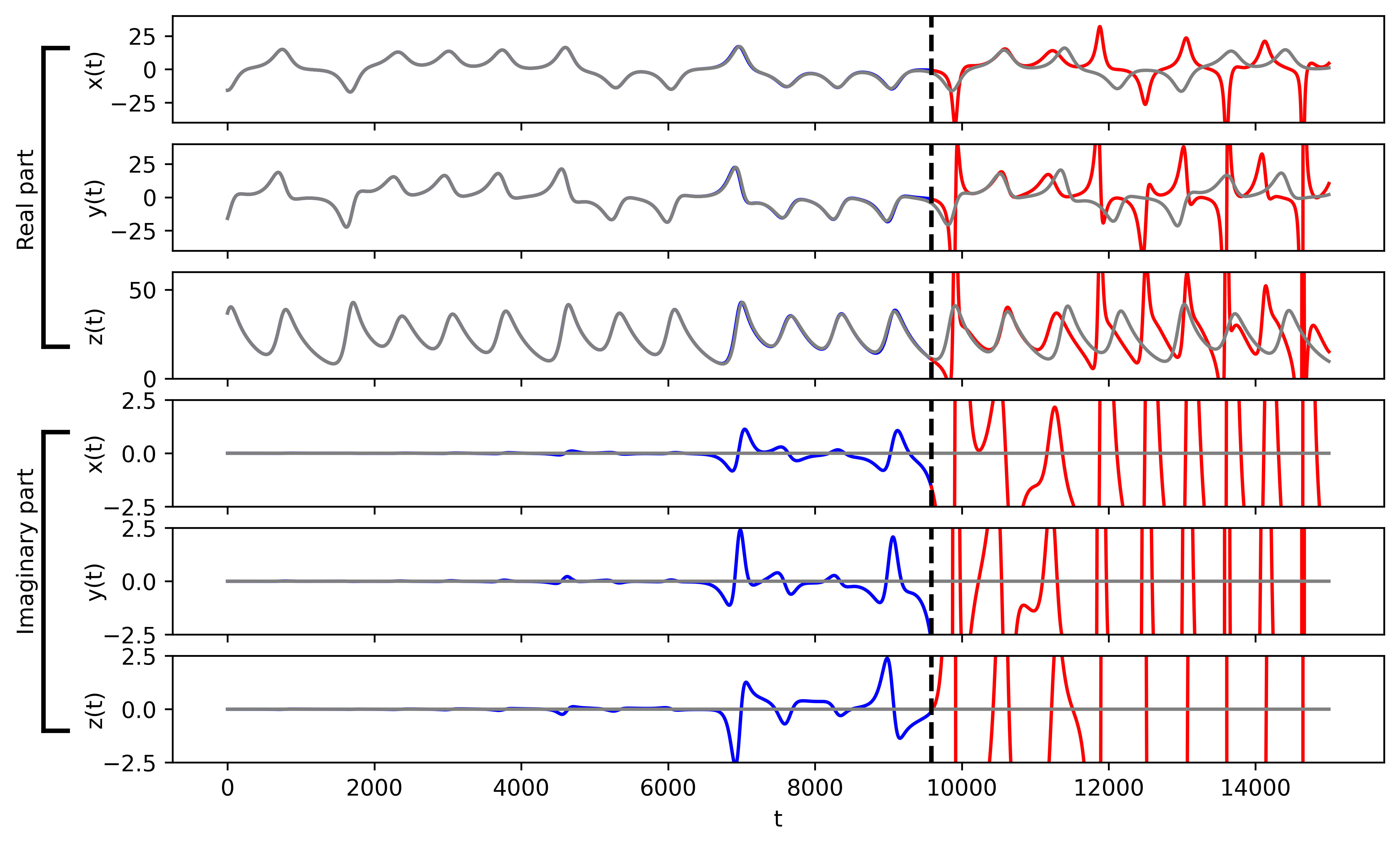}
    \end{subfigure}
    \hfill
    \begin{subfigure}[t]{0.32\textwidth}
        \caption{}
        \centering
        \includegraphics[width=5.95cm]{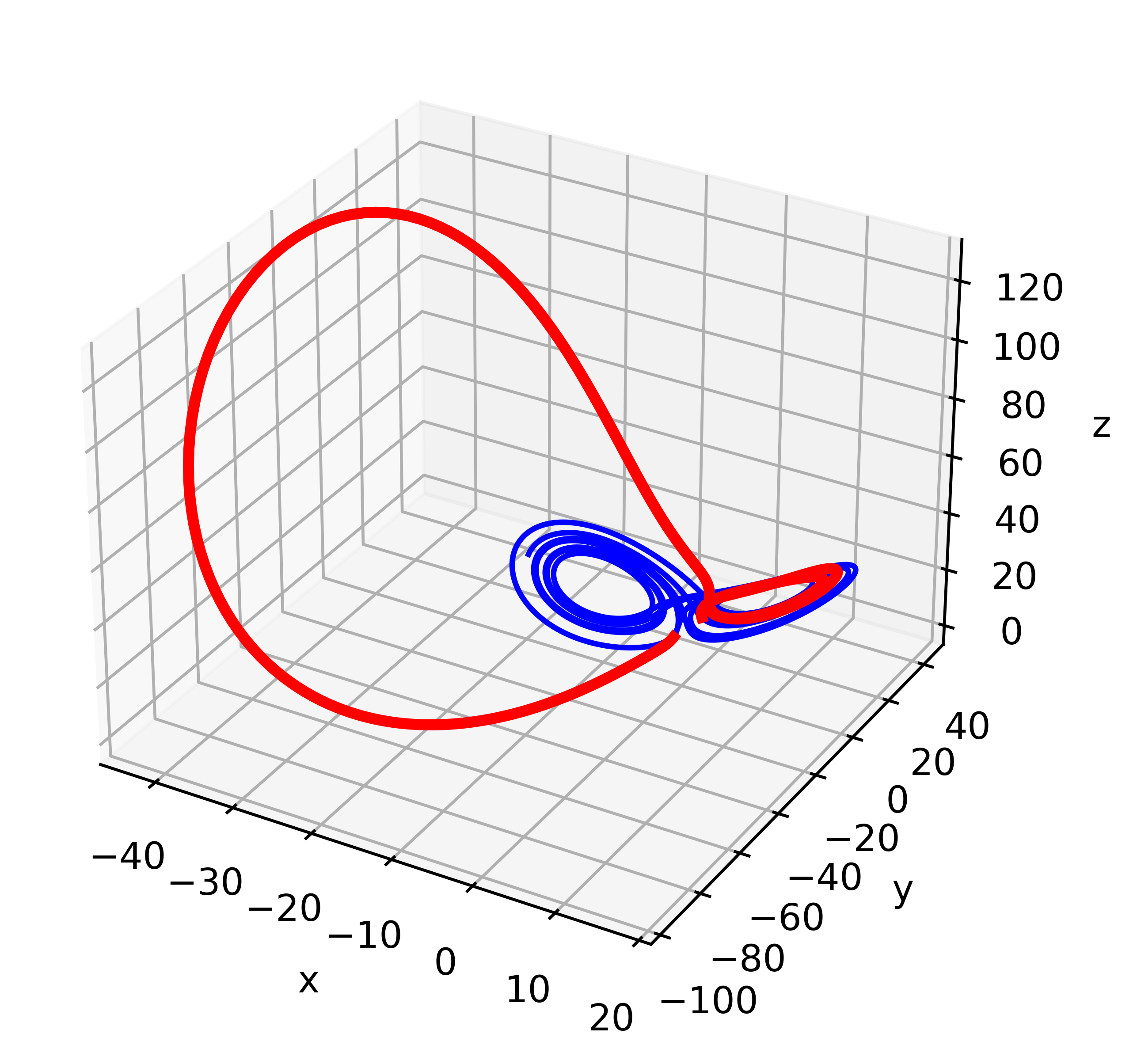}
        \vspace{1.25cm}
    \end{subfigure}

    \vspace{-1.5cm} 

    \begin{subfigure}[t]{0.62\textwidth}
        \caption{}
        \centering
        \includegraphics[width=11.05cm]{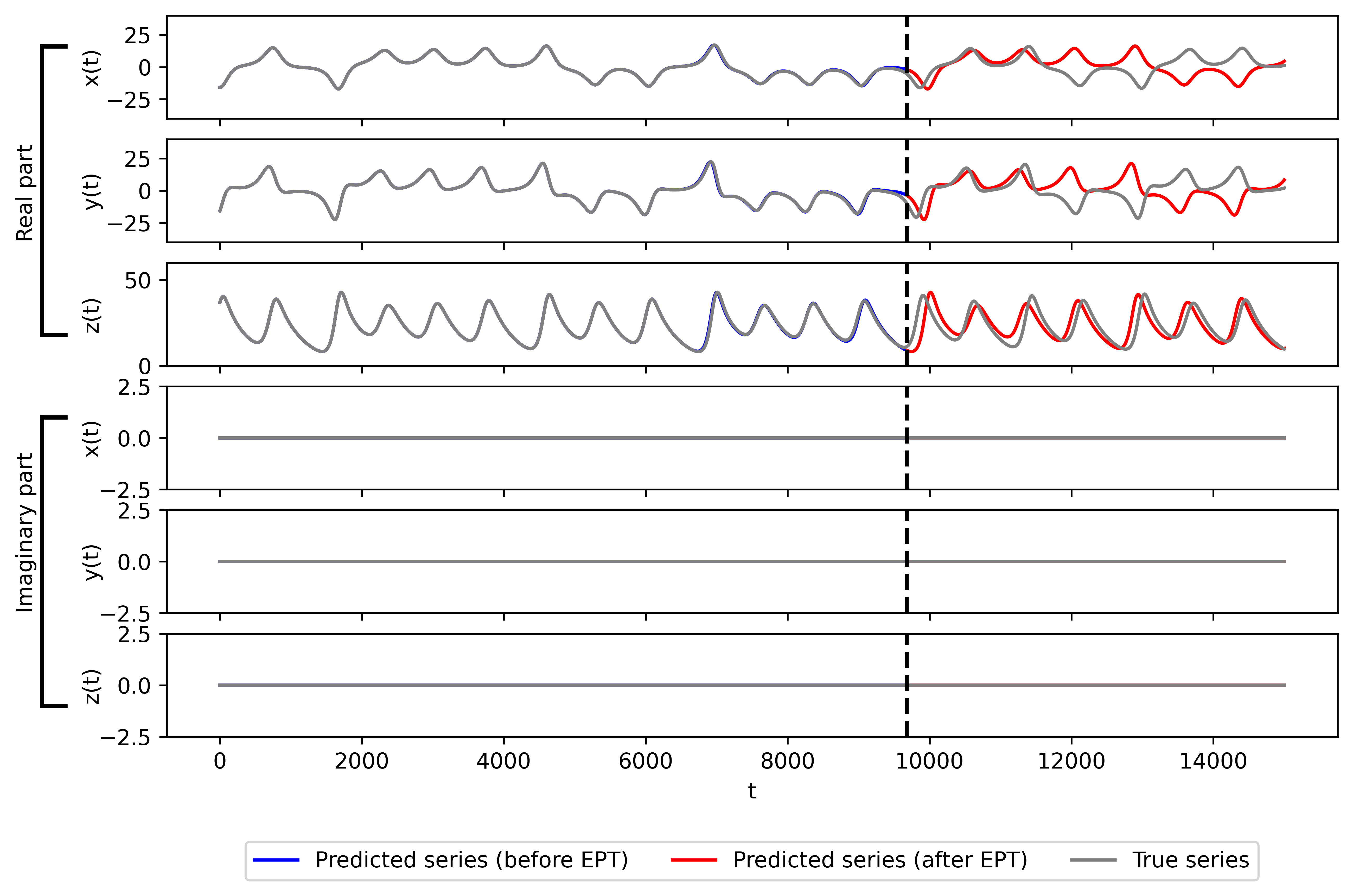}
    \end{subfigure}
    \hfill
    \begin{subfigure}[t]{0.32\textwidth}
        \caption{}
        \centering
        \includegraphics[width=5.95cm]{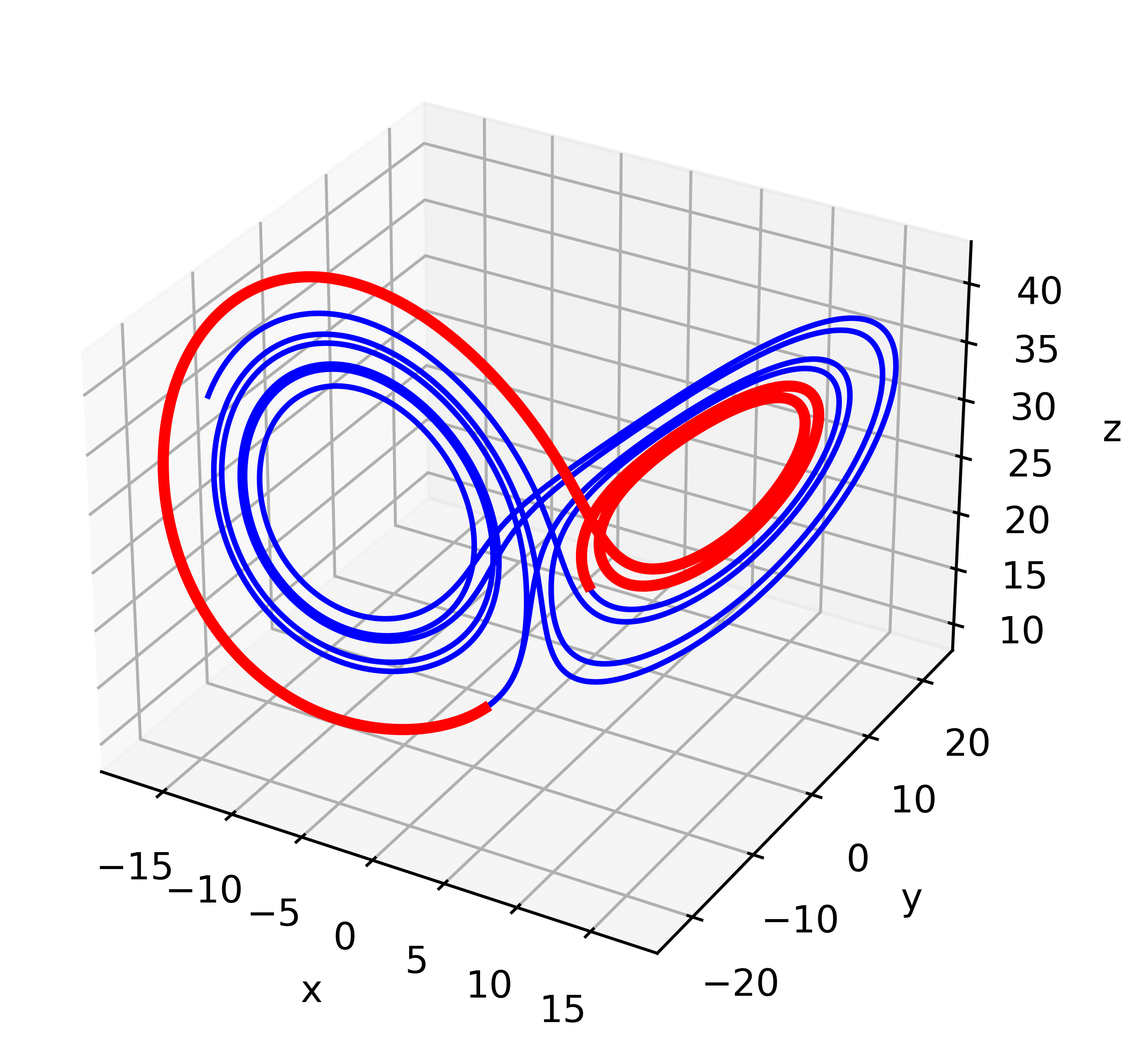}
        \vspace{1.25cm}
    \end{subfigure}
    \caption{Effect of erroneous model parameters on the dynamic behavior of
      the system. (a) Comparison of simulated trajectories obtained for the true
      Lorenz63 system (gray line) and a corrupted Lorenz63 system (blue and red
      lines), using identical starting values before and after the EPT are
      shown. After the $\widehat{EPT} = 8.68$ (vertical line), large deviations
      from the true trajectories are observed. Before the EPT at $9585$ points
      with $\widehat{EPT} = 8.68$, a growing imaginary part forebodes the EPT
      event. (b) The real component of the predicted series in phase space,
      distinguishing the trajectory before the EPT (blue line) and the $2000$
      points following the EPT (red line) is shown. (c) When omitting the
      imaginary part during time series prediction, the deviations become
      smaller and the EPT at $9684$ points with $\widehat{EPT} = 8.77$ improves
      by $99$ points. (d) The trajectories in phase space closely resemble the
      Lorenz attractor.}
    \label{fig:EPT_example}
\end{figure*}

The mean squared error is used for the comparison of two trajectories. The quadratic error term emphasizes large deviations at the level of individual coordinates. For complex-valued inputs, the mean squared error at time $t$ is defined as
\begin{equation}
E(t) = \frac{1}{n} \sum_{i=1}^{n} (X_i(t) - \tilde{X}_i(t))(\overline{X_i(t) - \tilde{X}_i(t)}).
\end{equation}
$X$ is the time series generated from the true system and $\tilde{X}$ denotes the time series produced by the model; $i$ refers to the respective coordinate or axis in space. 

The EPT is defined as the time elapsed before the prediction error, $E(t)$, exceeds a specified error threshold, $\theta$, for the first time. It can be formally expressed as follows:
\begin{equation}
EPT := \min\{t-1 ~| ~t \in \{1, \ldots, m\}, ~E(t) > \theta\}.
\end{equation}
The error threshold $\theta$ is chosen as the smallest standard deviation over all coordinates from the true series $X$.

For a dynamical system, contracting or expanding behavior can be expressed in terms of Lyapunov exponents. The largest Lyapunov exponent $\lambda_{\max}$ quantifies the degree of exponential divergence of close trajectories in the state space \cite{Pechuk2022}. The values obtained for the respective systems are shown in Table~\ref{table:Lyap}. 

\begin{table}[tbhp]
\centering
\begin{threeparttable}
\caption{Largest positive Lyapunov exponent per system.}
\label{table:Lyap}

\begin{tabular}{llll}
\toprule
Lorenz63\cite{Hurley2025} &
Four Wing\cite{Wang2009} &
Lorenz84\cite{Vannitsem2002} &
Lorenz\_Fract\tnote{a} \\
\midrule
$0.9056$ & $0.064$ & $0.56$ & $6 \times 10^{-5}$ \\
\bottomrule
\end{tabular}

\begin{tablenotes}
\footnotesize
\item[a] Calculated with the Python library \textit{Lyapynov} \cite{savary2025}.
\end{tablenotes}

\end{threeparttable}
\end{table}

For better comparison, we normalize the EPT by the Lyapunov Time, which is the reciprocal of this value $\frac{1}{\lambda_{\max}}$ \cite{Bezruchko2010}, and also multiply it by the step size $\Delta t$. Thus, we have
\begin{equation}
\widehat{EPT} := EPT \times \Delta t \times \lambda_{\max}.
\end{equation}

\vspace{\baselineskip}
Fig.~\ref{fig:EPT_example} shows an example of the EPT. A trajectory generated by the true Lorenz63 system is compared with one evolved for corrupted Lorenz63 system in which the first equation is altered as
\begin{equation}
    \dot{x} = -10x + (10.001+0.001i)y,
\end{equation}
resembling some inferences produced by our product-unit model. 

Even after rounding to three decimal places, small residual imaginary parts may remain. These components also contribute to the prediction error and reduce the effective prediction time of the trajectory. Moreover, through nonlinear couplings, imaginary parts can also influence the real components of the equations. Their removal from the system after training can thus improve the stability of the trajectory evolution, as can be seen in Fig.~\ref{fig:EPT_example}. However, removing the imaginary component from the system did not substantially increase the EPT (from $8.68$ to $8.77$), a behavior also observed for most systems inferred by the product-unit model. 

We frequently observed that changes in the imaginary parts serve as a reliable early indicator of an imminent deviation from the expected trajectory (e.g., Fig.~\ref{fig:EPT_example}a). In a previous study on nuclear mass predictions with complex product-unit networks, the imaginary component of the output was assumed to indicate prediction uncertainty \cite{DellenEtAl2024}. For this reason, and because the EPT generally did not show significant increases without them, we do not discard the imaginary parts, aside from rounding to the third decimal place, in the simulations.

\section{Results}
We evaluate the model on four benchmark systems and subsequently apply it to real-world gait data, assessing both symbolic recovery accuracy and predictive stability.
For the benchmark, we use the well-known Lorenz63 system \cite{Lorenz1963}, Lorenz84 \cite{Vannitsem2002} and Four Wing \cite{Wang2009}, each described by three coupled nonlinear ordinary differential equations. In addition, we examine a modified version of Lorenz63, Lorenz\_Fract, which contains a fractional exponent. Throughout the benchmark experiments, derivatives are treated as directly observable quantities rather than being estimated from discretely sampled states. For the gait application, the problem is substantially higher-dimensional; here, the model predicts the next state solely from a set of lagged previous observations, as analytic equations are unavailable.

The loss function is taken as the mean squared error in the following, defined for complex-valued inputs by
\begin{align}\label{eq:MSE}
CMSE = \frac{1}{n} \sum_{i=1}^{n} &(f_{i} - \hat f_{i}) \times \overline{(f_{i} - \hat f_{i})},
\end{align}
so that $f_{i}$ denotes the true function value, while $\hat f_{i}$ represents the prediction produced by the product-unit model. For all the dynamical systems considered, there are three state variables ($n=3$).

\subsection{Performance on Benchmark Systems}\label{section:results_benchmark}
The training data is prepared as outlined in Sec.~\ref{section:data}, for different choices of points and trajectories. The expected output for the points along a trajectory is obtained by evaluating the system's equations at those points. This output, $\dot{x}_i := f_{i}(x_1, \ldots ,x_n)$, is then compared to the model's predicted results, $\hat f_{i}(x_1, \ldots ,x_n)$, within the loss function (Eq.~\ref{eq:MSE}).

The parameters are kept the same across all systems, except for two changes in the generation of training data for Lorenz\_Fract (Sec.~\ref{section:data}).

Training is performed for $5000$ epochs using the Adam algorithm \cite{kingma2017}. The initial learning rate is set to $0.03$ for the coefficients and to $0.003$ for the weights of the exponents, with both rates exponentially decayed by a factor of $\gamma = 0.999$ at each epoch. The batch size is set to $30$. The model uses the minimum number of product units needed to accurately represent the system. For Lorenz63 and Lorenz\_Fract, this is five; for Four Wing, six; and for Lorenz84, eight.

We count the final number of discovered terms and subsequently measure the \textit{effective prediction time} (EPT) of the resulting system (Sec.~\ref{section:modelevaluation}).

\begin{figure}[htbp]
  \centering
  \begin{subfigure}[b]{1.\linewidth}
    \centering
    \includegraphics[width=8.5cm]{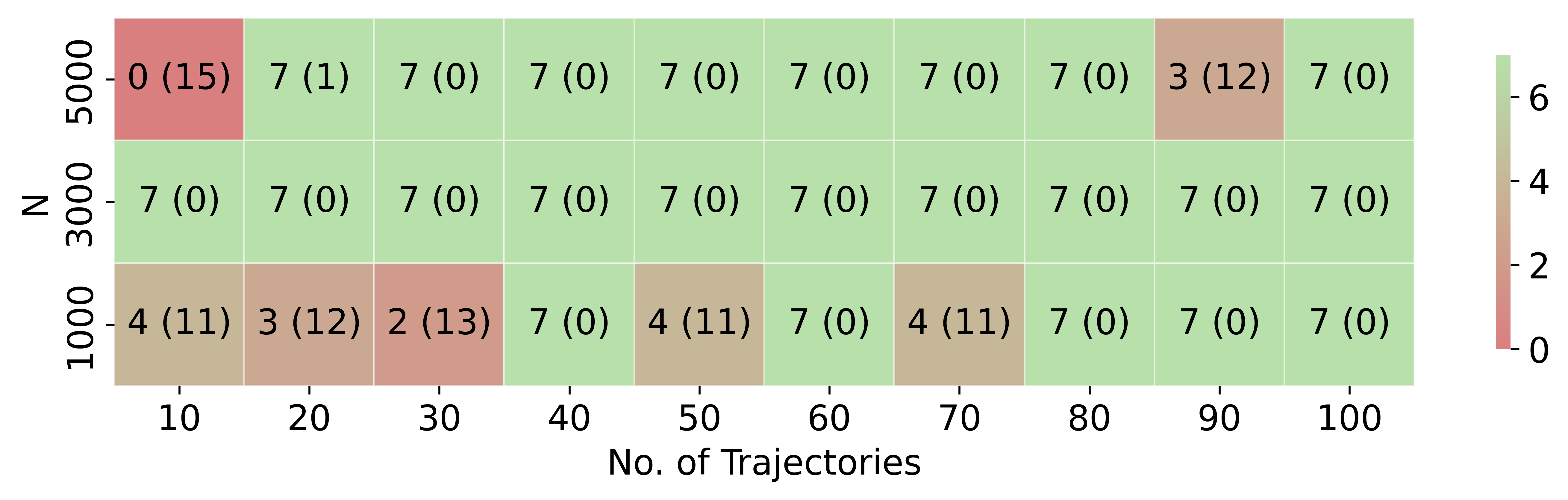}
    \caption{Lorenz63, five product units.}
  \end{subfigure}
  \begin{subfigure}[b]{1.\linewidth}
    \centering
    \includegraphics[width=8.5cm]{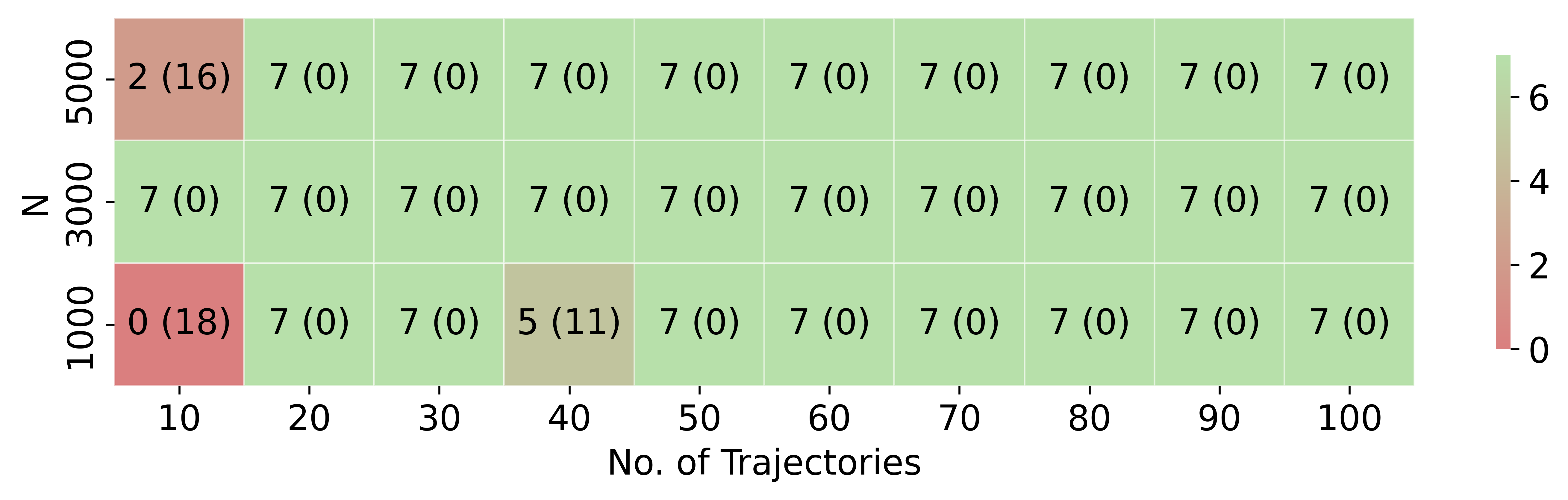}
    \caption{Four Wing, six product units.}
  \end{subfigure}
  \begin{subfigure}[b]{1.\linewidth}
    \centering
    \includegraphics[width=8.5cm]{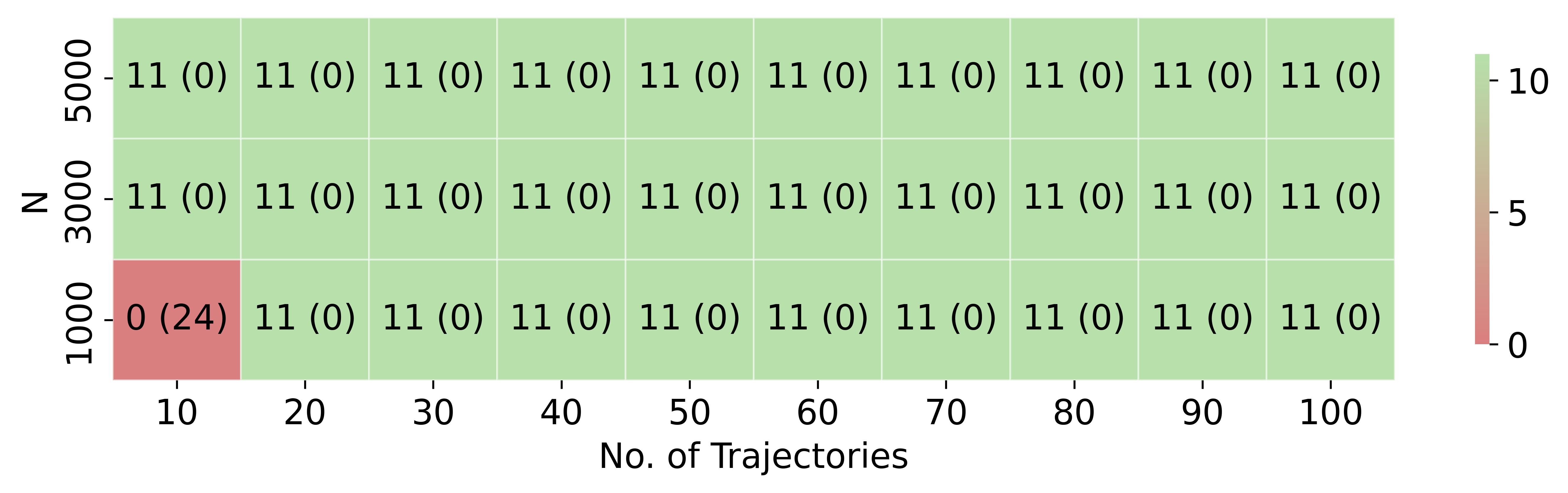}
    \caption{Lorenz84, eight product units.}
  \end{subfigure}
  \begin{subfigure}[b]{1.\linewidth}
    \centering
    \includegraphics[width=8.5cm]{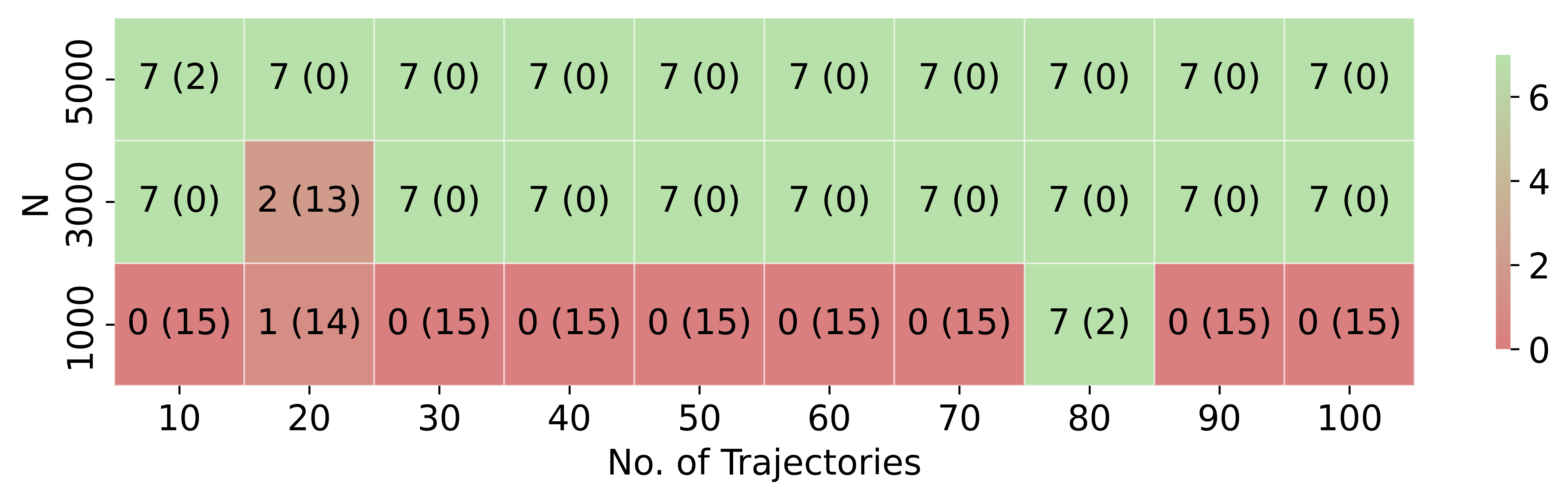}
    \caption{Lorenz\_Fract, five product units.}
  \end{subfigure}
  \caption{Heatmap of the number of correctly identified terms across different combinations of points and trajectories. The number of erroneous terms is indicated in brackets. For illustration, the minimum number of product units required is used for each system. When at least $3000$ points and their derivatives are supplied for training, the model consistently identifies the precise terms and rarely introduces spurious ones.}
  \label{fig:Terms}
\end{figure}

\subsubsection{Model discovery}

We repeatedly observed a significant relationship between the number of correctly identified and erroneous terms. When not all terms were discovered, the number of erroneous terms tended to be higher. Conversely, in cases where all expected system terms were correctly identified---except for three instances in total---no erroneous terms were observed.

In Fig.~\ref{fig:Terms}, the number of terms correctly recovered by the product-unit model is shown for the different dynamical systems and choices of points and trajectories. The number of terms identified as erroneous is reported in brackets.

The product-unit model demonstrated strong performance in representing the functions of dynamical systems. For Lorenz63, Four Wing and Lorenz84, all terms of the respective equations were correctly recovered in about $88\%$ of the trials, across varying numbers of points and trajectories. When at least $3000$ points were used for training, this proportion increased to $95\%$.

With the modifications to the training data outlined in Sec.~\ref{section:data}, similar results to those of the other systems were obtained for Lorenz\_Fract. While $1000$ points were rarely sufficient to learn the equations, $3000$ points already resulted in complete representations in $9$ out of $10$ trials.

Apart from the choice of the optimizer or learning rate, the most relevant and challenging parameter to determine is the number of product units, which must be equal to or greater than the number of distinct monomials in the system. Note that the same units appear in each equation, differing only in their coefficients (Sec.~\ref{section:model}). To obtain a minimal representation, we permit merging of terms that are sufficiently similar to one another (Sec.~\ref{section:modelevaluation}).

To estimate the influence of more available product units, we repeated the analysis depicted in Fig.~\ref{fig:Terms} for Lorenz63 with varying choices of this parameter $m$ in Eq.~\ref{eq:2}. A larger number of product units may offer more opportunities to correctly identify the terms, but it also increases model complexity. 

\begin{figure}[htbp]
  \centering
  \begin{subfigure}[b]{1.\linewidth}
    \centering
    \includegraphics[width=8.5cm]{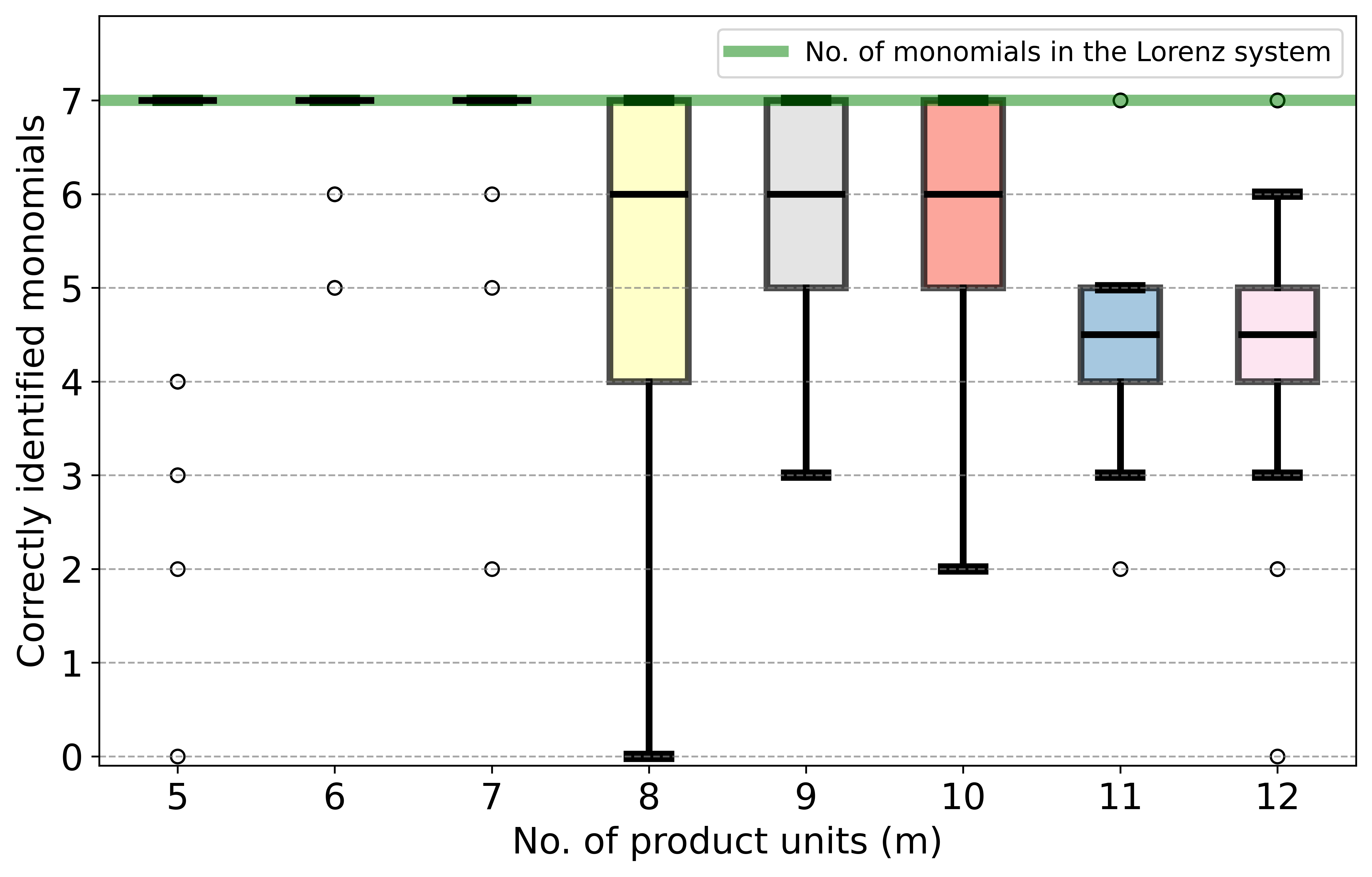}
    \caption{}
  \end{subfigure}
  \begin{subfigure}[b]{1.\linewidth}
    \centering
    \includegraphics[width=8.5cm]{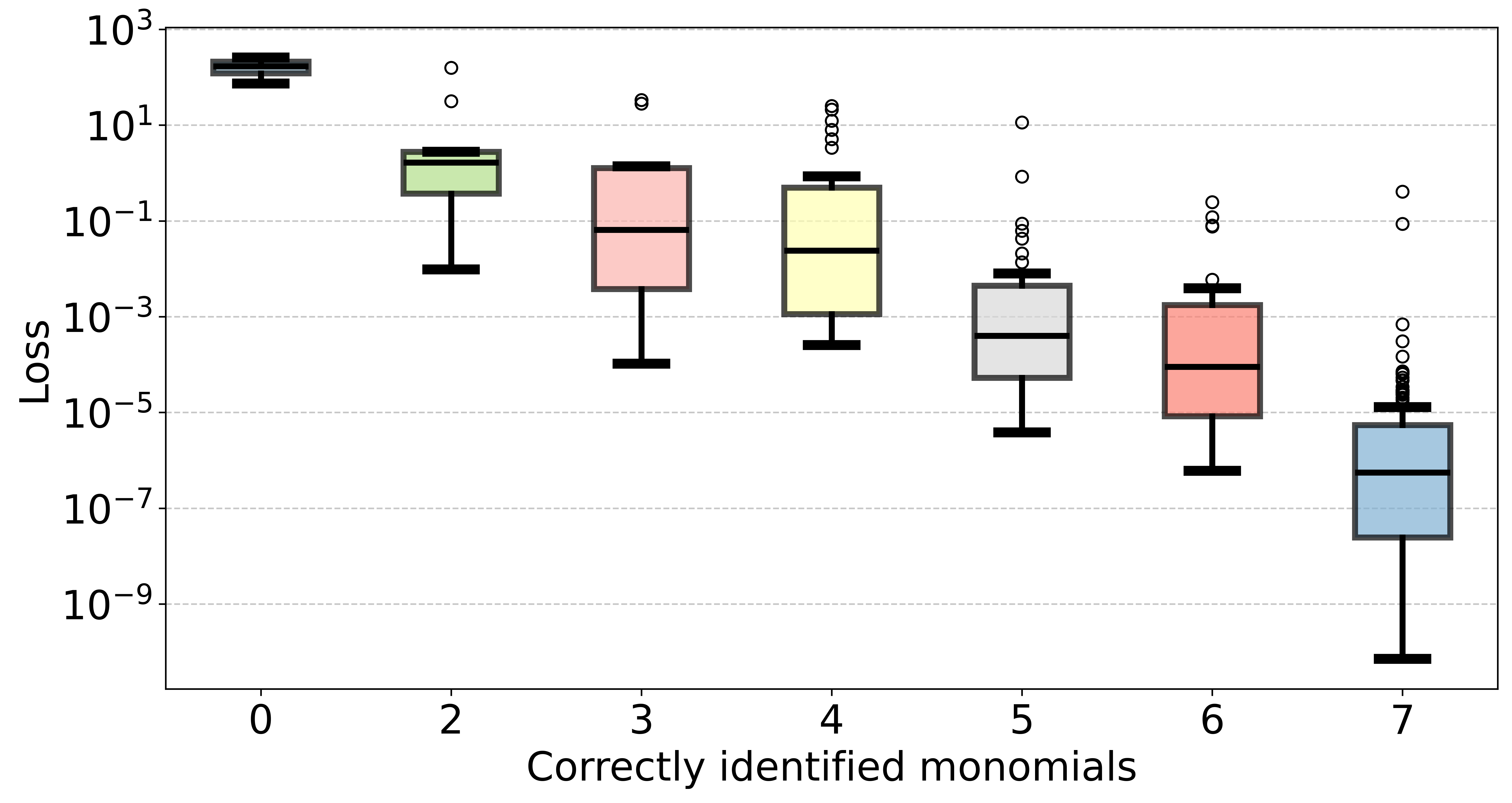}
    \caption{}
  \end{subfigure}
  \caption{Dependence of the number of correctly identified terms in the Lorenz63 model on the number of product units in the network (a), and dependence of the test loss on the number of correctly identified terms (b). For $5-7$ product units, all terms are correctly identified in at least $75\%$ of cases, and for $8-10$ product units in at least $25\%$ of cases (a). The test loss is indicative of the number of correctly identified terms (b).}
  \label{fig:TermsPUs}
\end{figure}

Figure~\ref{fig:TermsPUs}a indicates that employing slightly more product units than minimally required for term representation does not significantly alter the model's ability to accurately identify the expressions. For the Lorenz63 system, there are seven terms in total, which can be represented with five product units. Best performance was achieved with six product units, but overall five to seven units produced similar results. With ten units, full recovery of all terms is still achieved in at least $25\%$ of cases.

Furthermore, when the underlying dynamics are unknown and there is no prior expectation regarding the number of expressions involved, Figure~\ref{fig:TermsPUs}b suggests that the loss can provide an indication of the number of dominant terms in the system.




\subsubsection{Time series prediction}

Prediction of chaotic time series is often done using deep neural networks, which generally demonstrate strong performance \cite{Vlachas2018, Chattopadhyay2020, Choi2022, Chen2020}. In a recent work using reservoir computing, more than $30$ Lyapunov times could be predicted for the Lorenz system \cite{Hurley2025}. However, the black-box character of many of these approaches severely limits the interpretability of the recovered models. Moreover, extracting some of the key monomials may also improve forecasting performance, motivating the application of our method to this task \cite{Wang2024}.

\begin{figure}[htbp]
  \centering
  \begin{subfigure}[b]{\linewidth}
    \centering
    \includegraphics[width=8.5cm]{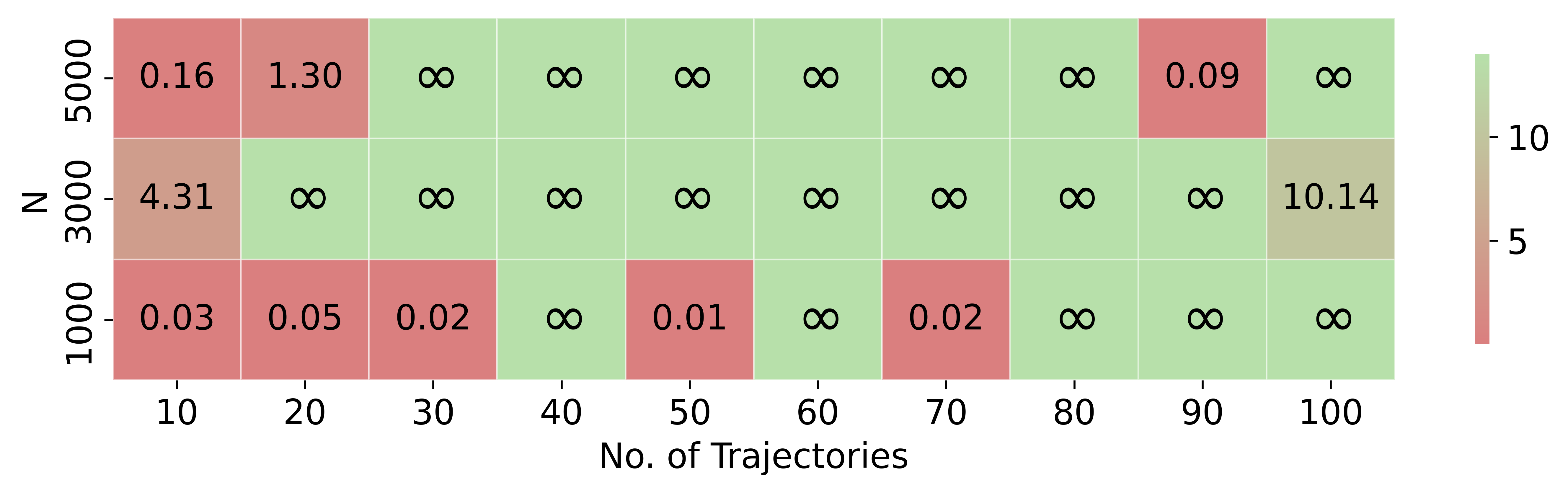}
    \caption{Lorenz63, five product units.}
  \end{subfigure}
  \begin{subfigure}[b]{1.\linewidth}
    \centering
    \includegraphics[width=8.5cm]{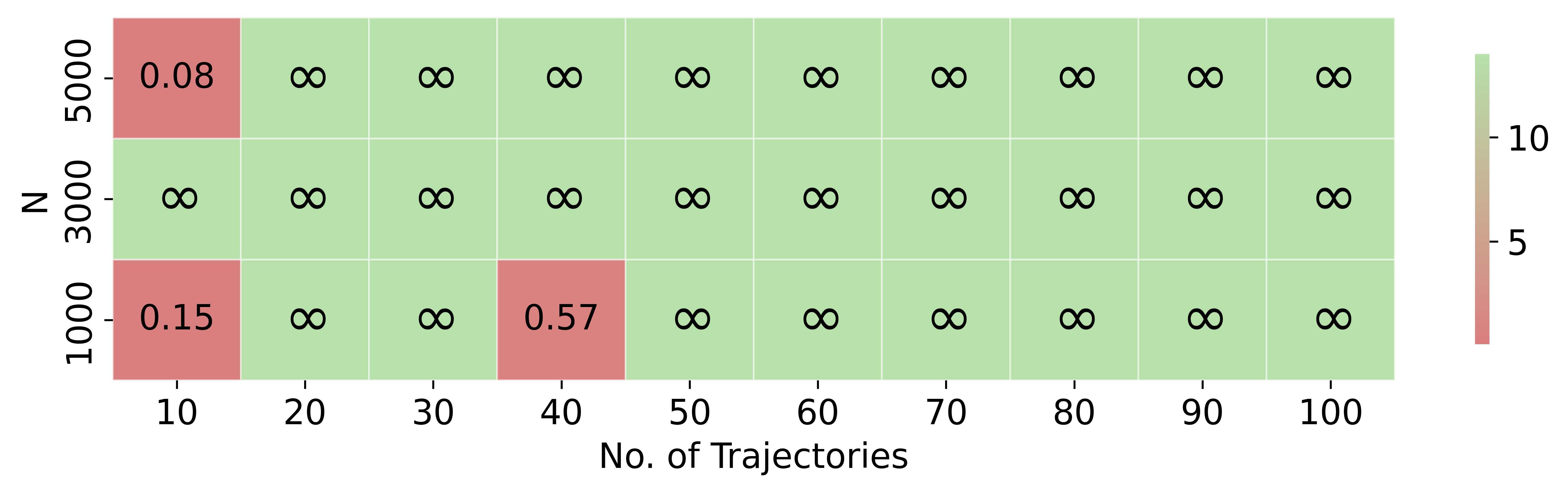}
    \caption{Four Wing, six product units.}
  \end{subfigure}
  \begin{subfigure}[b]{1.\linewidth}
    \centering
    \includegraphics[width=8.5cm]{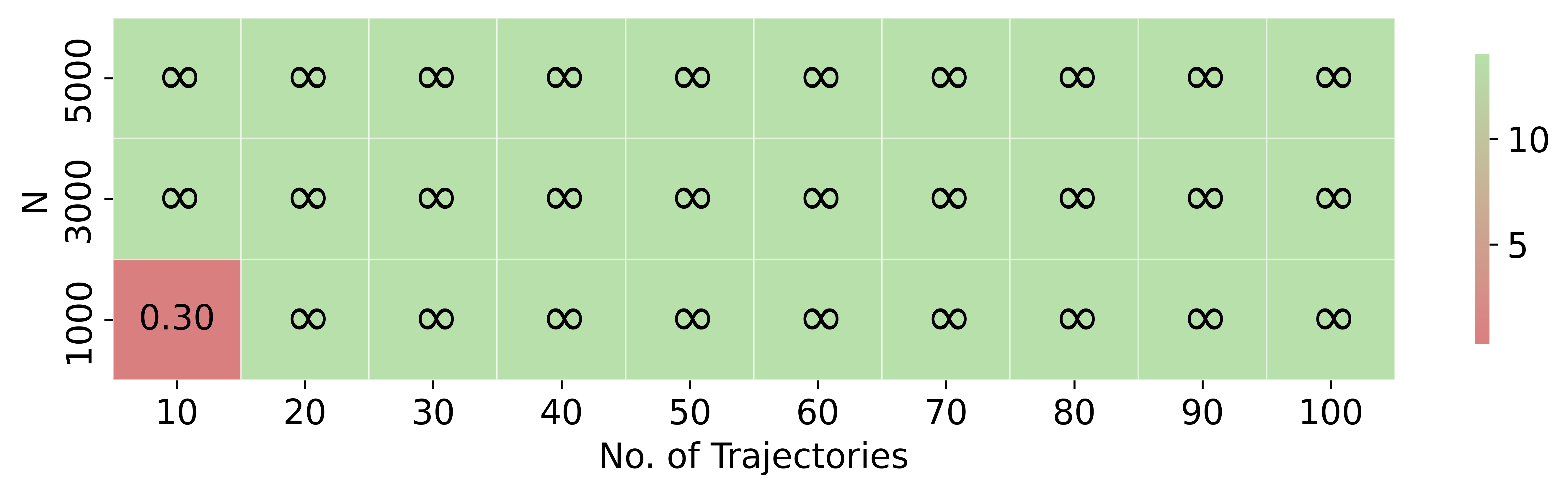}
    \caption{Lorenz84, eight product units.}
  \end{subfigure}
  \begin{subfigure}[b]{1.\linewidth}
    \centering
    \includegraphics[width=8.5cm]{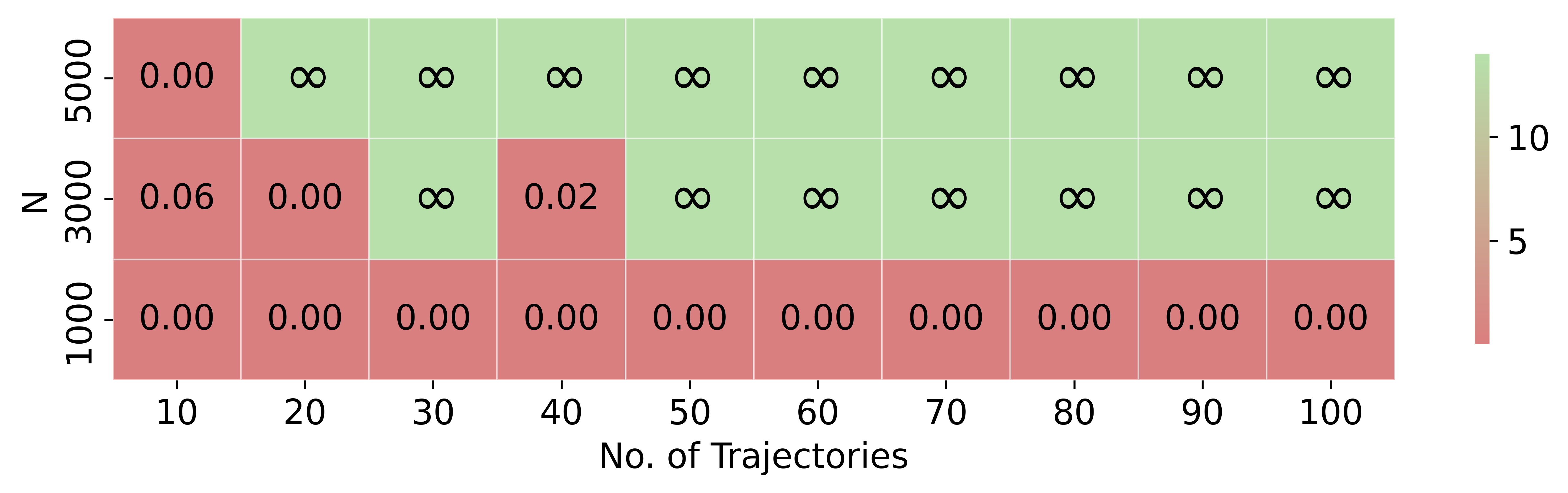}
    \caption{Lorenz\_Fract, five product units.}
  \end{subfigure}
  \caption{Heatmap of the effective prediction time for systems recovered by the product-unit model under varying training data. A high $\widehat{EPT}$ indicates stable predictive performance. Rounding to the third decimal point is performed in advance, and this often results in the discovery of the exact equations of the system. In those cases, $\widehat{EPT} = \infty$, as the predicted and expected systems are indistinguishable.}
  \label{fig:EPT}
\end{figure}

The effective prediction time (EPT) is calculated as described in Sec.~\ref{section:modelevaluation}. The EPT measures the elapsed time before the predicted trajectory significantly deviates from the trajectory with the same initial coordinates, but governed by the true system. As such, a higher EPT signifies more stable predictive performance. Fig.~\ref{fig:EPT} shows the effective prediction time for the different dynamical systems and choices of points and trajectories. Even when all system terms are identified by our counting method, small deviations from the true equations can remain, resulting in a finite EPT.

For the product-unit model, the EPT on the benchmark systems highly depends on its ability to resolve all of the leading terms. When it failed to identify at least one of them, the resulting EPT consistently fell below one. However, with a sufficient amount of training data, the produced equations were often identical to expectation, allowing the underlying dynamics to be predicted indefinitely. The exact systems of Lorenz63, Lorenz84 and Four Wing were resolved in $84\%$ of all trials and in $90\%$ of experiments with at least $3000$ points, although a slight difference in success was observed between Lorenz63 and the other two systems of ODEs (Fig.~\ref{fig:EPT}a-c). Using $3000$ training points for Lorenz\_Fract produced models identical to the true system in seven out of ten cases, and $5000$ points increased this to nine out of ten (Fig.~\ref{fig:EPT}d). 


\subsection{Application to Human Gait Signals}
To demonstrate the model’s performance on a high-dimensional system, the
application to human gait analysis is explored. The problem involves a finite
walking pattern signal consisting of acceleration values in three directions,
with the goal of forecasting the time series behavior.

The numerous variables involved in human gait make it difficult to derive
precise future predictions, but we aim to produce a stable trajectory that
approximates the true signal.

\begin{figure*}[htbp]
    \centering
    \begin{subfigure}[b]{1.0\textwidth}
        \centering
        \includegraphics[width=\textwidth]{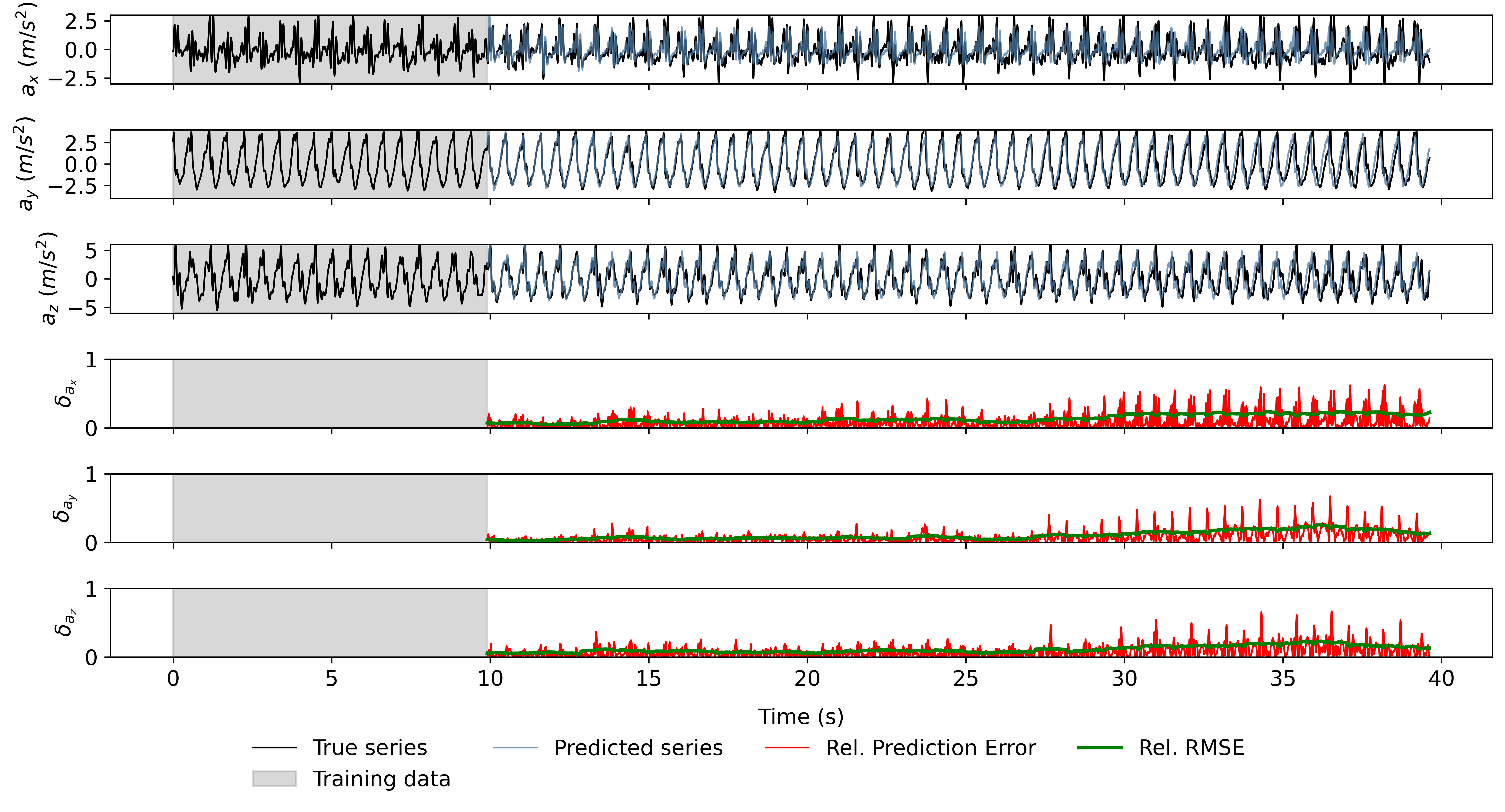}
        \caption{}
    \end{subfigure}


    \begin{subfigure}[b]{0.55\textwidth}
        \centering
        \includegraphics[width=\textwidth]{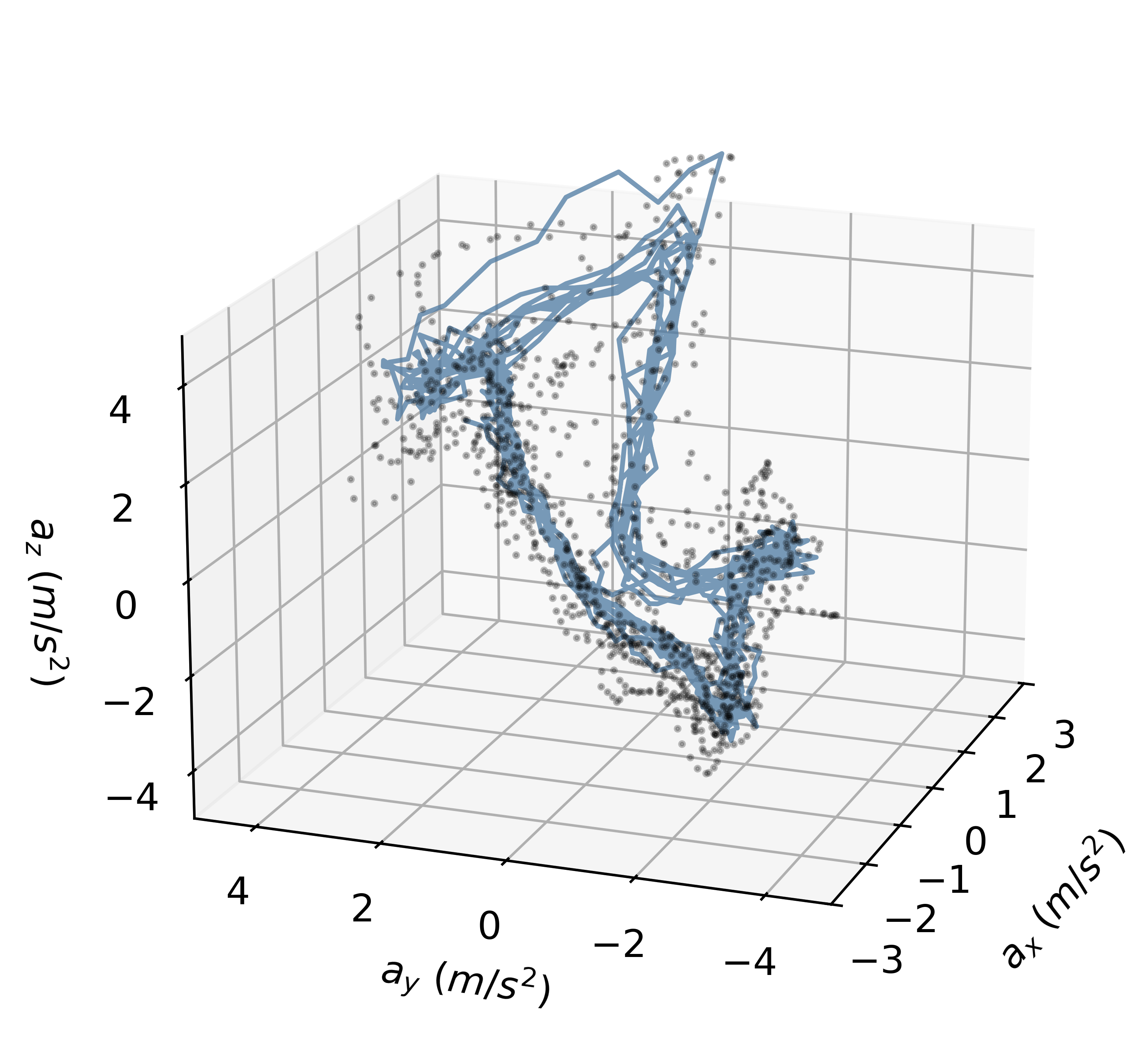}
        \caption{}
    \end{subfigure}
    \caption{Human walking acceleration signals predicted by the product-unit
      model. (a) True signal trajectory (black) and product-unit model output
      (steel-blue) in the $x$, $y$, and $z$ directions. Data from the first ten
      seconds are used for training, as indicated by the dashed black line. The
      inferred system precisely predicts the further course of $a_y$ and $a_z$
      for the considered time frame. The lower three panels show the relative prediction error $\delta_{a_v}(t) = \frac{|a_v^{\rm true}(t) - a_v^{\rm pred}(t)|}{\max_t a_v(t) - \min_t a_v(t)}$ for each direction, normalized by the signal peak-to-peak range, i.e. the difference between its maximum and minimum values. The error remains bounded throughout the 30\,s test interval, with a small upward trend, and inherits the periodic structure of the gait cycle. The green line in each error panel marks the moving root-mean-square error (RMSE), computed over a sliding window of approximately one second and normalized by the signal peak-to-peak range. (b) The first five seconds of the two trajectories in the 3D phase space. Strong quasi-periodic behavior is observed, as expected for a steady walking cycle.}
    \label{fig:Acc}
\end{figure*}

As described in Sec.~\ref{section:data}, we consider ten walking signals, each
containing $8000$ points, of which $2000$ are used for training. In contrast to
the benchmark instances, the next state in the acceleration time series along
the $x,y$ and $z$ directions is modeled as a function of $50$ previous states
sampled at various time lags $\psi_k$. In the loss function described earlier
(Eq.~\ref{eq:MSE}), the prediction of the product-unit model,
$\hat f_i(a(t-\psi))$, is thus compared against the true coordinates $a_i(t)$ of
the next point.

$300$ product units are used for the representation. Training is performed for
$500$ epochs with a batch size of $30$ using the Adam algorithm. The initial
learning rate is set to $0.03$ for the coefficients and $0.003$ for the exponent
weights, with the rates decayed by $\gamma=0.99$ per epoch.

The product-unit model routinely captures the dynamics of the corresponding
signals: Within only a few hundred epochs, the model consistently converges to a
stable trajectory that provides a reasonable projection based on the training
data and closely resembles the true signal for at least a few thousand data
points (e.g., see Fig.~\ref{fig:Acc}), with one second corresponding to
approximately $200$ points. 
In some cases, the signal closely matched the expected trajectory throughout the entire time interval, although $a_x$ proved slightly more difficult to learn due to its more complex dynamics compared to the other directions.



Quantitatively, the relative prediction error remained bounded throughout the 30\,s test interval. The root-mean-square error (RMSE) is computed over the full signal and normalized by its amplitude range, defined as the difference between the maximum and minimum values
\[
\widehat{\mathrm{RMSE}}_v = \frac{\mathrm{RMSE}_v}{\max_t a_v(t) - \min_t a_v(t)},
\]
yielding a relative RMSE of approximately $\widehat{\mathrm{RMSE}}_x \approx 0.14$, $\widehat{\mathrm{RMSE}}_y \approx 0.12$, and $\widehat{\mathrm{RMSE}}_z \approx 0.12$. 
In Fig.~\ref{fig:Acc}, a \textit{moving} RMSE is shown, computed over a one-second sliding window and also normalized by the full signal amplitude range. Importantly, only moderate growth of the error over time was observed, indicating that the model does not accumulate substantial drift on a horizon three times longer than the training interval.

\section{Summary and Conclusions}

This work addresses the problem of recovering the governing equations of
dynamical systems directly from observed trajectory data. We proposed the
complex-valued product-unit model as a library-free approach to sparse dynamical
system representation, in which the relevant monomials, including those with
fractional or negative exponents, are learned directly from data rather than
selected from a predefined candidate set. This distinguishes the approach from
established methods such as SINDy or symbolic regression, which rely on a fixed
function library, while preserving the potential to obtain sparse and
interpretable representations of the underlying dynamics.

The approach was evaluated on four benchmark systems of coupled ordinary
differential equations: Lorenz63, Lorenz84, Four Wing, and a fractional variant
of Lorenz63 (Lorenz\_Fract). In the majority of trials, the product-unit model
reliably recovered the precise terms of each system. When at least 3000 training
points were provided, the exact equations were identified in 90\% of experiments
for the integer-exponent systems, and in 70–90\% of trials for Lorenz\_Fract,
demonstrating that the approach extends naturally to non-integer power laws.
This is of particular importance for high-dimensional problems that may consist
of many different non-linear interactions, making it difficult to predefine
relevant monomials before training. The recovered models, assessed via the
effective prediction time (EPT), remained stable and close to the well-known
attractors even beyond the EPT when the spurious imaginary terms were discarded.
On the other hand, exploding imaginary components were observed close to the EPT
and support earlier observations that the emergence of spurious imaginary
components may serve as a useful indicator of the reliability of the predictions
of product-unit networks.



The method was subsequently applied to real-world human gait signals — a
substantially higher-dimensional problem in which analytic governing equations
are unavailable. Rather than learning a differential equation, the model was
trained in this case to predict the next state from a series of previous
observations. Using a time-delay embedding formulation, the model produced
stable and plausible trajectories from short training sequences, with bounded, non-divergent prediction errors over a test horizon three times longer than the training interval, demonstrating
its applicability beyond controlled benchmark settings.


Several limitations of the current approach deserve attention. First, the number
of product units must be specified in advance and must be at least equal to the
number of dominant terms in the system. While this is a non-trivial requirement
when the system structure is unknown, Fig.~\ref{fig:TermsPUs}b shows that the training loss
provides some indication of model adequacy. Second, the interpretability of the
model degrades rapidly with dimensionality: the gait application required 300
product units with 150 exponents each, yielding about 46,000 learnable
parameters — a regime in which the sparse, human-readable representation that is
the primary motivation of model discovery is largely lost. Regularization
strategies and systematic term-merging procedures may help recover sparsity in
such settings. Third, the model's performance on noisy data has not been
systematically evaluated; the gait signals required low-pass filtering as a
preprocessing step, and robustness to measurement noise remains to be studied
more carefully.

Future work could address these limitations in several directions. Adaptive or
penalized methods for determining the number of product units automatically
would broaden the practical applicability of the approach. Extending the
framework to partial differential equations, in the spirit of recent work on
physics-informed learning, is another natural direction. Finally, a more
systematic evaluation of the gait application, including quantitative metrics
across multiple subjects and walking conditions, may open pathways for
deployment in biomedical contexts. Relevant domains of application include the
detection of near-fall situations or gait anomalies, e.g.\ connected to
neurodegenerative diseases. In this setting, the straightforward implementation,
the low resource requirements, and the relatively small set of hyperparameters
allow training and inference on edge computing devices, an important
consideration in privacy-sensitive applications.

\bibliographystyle{unsrt}  
\bibliography{dynamics}

\end{document}